\documentclass[10pt,twocolumn,letterpaper]{article}

\usepackage{cvpr}              

\usepackage{mathtools}
\usepackage{bm}
\usepackage{pifont}
\usepackage{multirow}
\usepackage{xcolor}
\usepackage{stfloats}
\definecolor{cvprblue}{rgb}{0.21,0.49,0.74}
\usepackage[pagebackref,breaklinks,colorlinks,allcolors=cvprblue]{hyperref}

\def\paperID{-} 
\def\confName{CVPR}
\def\confYear{2025}

\title{X as Supervision: Contending with Depth Ambiguity in Unsupervised Monocular 3D Pose Estimation}

\author{Yuchen Yang$^{1,3}$\thanks{Work performed during his internship at Shanghai Artificial Intelligence Laboratory.} 
\ \
Xuanyi Liu$^{2,3}$
\ \
Xing Gao$^{3}$
\ \
Zhihang Zhong$^{3}$
\  \
Xiao Sun$^{3}$\thanks{Corresponding arthor} \\
$^1$Fudan University \ \
$^2$Peking University \ \
$^3$Shanghai Artificial Intelligence Laboratory \\
{\tt\small yangyc22@m.fudan.edu.cn \ \ xuanyi@stu.pku.edu.cn \ \{gaoxing, zhongzhihang, sunxiao\}@pjlab.org.cn}
}

\begin{document}
\maketitle
\begin{abstract}
Recent unsupervised methods for monocular 3D pose estimation have endeavored to reduce dependence on limited annotated 3D data, but most are solely formulated in 2D space, overlooking the inherent depth ambiguity issue.
Due to the information loss in 3D-to-2D projection, multiple potential depths may exist, yet only some of them are plausible in human structure.
To tackle depth ambiguity, we propose a novel unsupervised framework featuring a multi-hypothesis detector and multiple tailored pretext tasks.
The detector extracts multiple hypotheses from a heatmap within a local window, effectively managing the multi-solution problem.
Furthermore, the pretext tasks harness 3D human priors from the SMPL model to regularize the solution space of pose estimation, aligning it with the empirical distribution of 3D human structures.
This regularization is partially achieved through a GCN-based discriminator within the discriminative learning, and is further complemented with synthetic images through rendering, ensuring plausible estimations.
Consequently, our approach demonstrates state-of-the-art unsupervised 3D pose estimation performance on various human datasets. 
Further evaluations on data scale-up and one animal dataset highlight its generalization capabilities. 
Code will be available at \url{https://github.com/Charrrrrlie/X-as-Supervision}.
\end{abstract}    
\section{Introduction}
3D pose estimation has shown great potential in various domains, including robotics~\cite{cheng2024expressive,luo2023perpetual}, augmented/virtual reality~\cite{chen2024within}, and human behavior analysis~\cite{yan2018spatial}, where monocular system offers significant advantages in terms of streamlined deployment.
However, accurate 3D annotations for supervised training of 3D pose estimation systems come from motion capture environments, which are hard and costly to obtain.
It relates to multi-view cameras~\cite{ionescu2013human3,mono-3dhp2017} or assistive sensors (IMUs, etc)~\cite{vonMarcard2018,shahroudy2016ntu}, involving labor-intensive and costly setups and processes, which restricts data collection scale, thereby hampering the ability of 3D pose estimation systems.
These limitations inspire the exploration of unsupervised 3D pose estimation, aiming to reduce the reliance on manual annotations and leverage vast in-the-wild data.

\begin{figure*}[th]
    \begin{subfigure}[b]{0.45\linewidth}
        \centering
        \includegraphics[width=\textwidth]{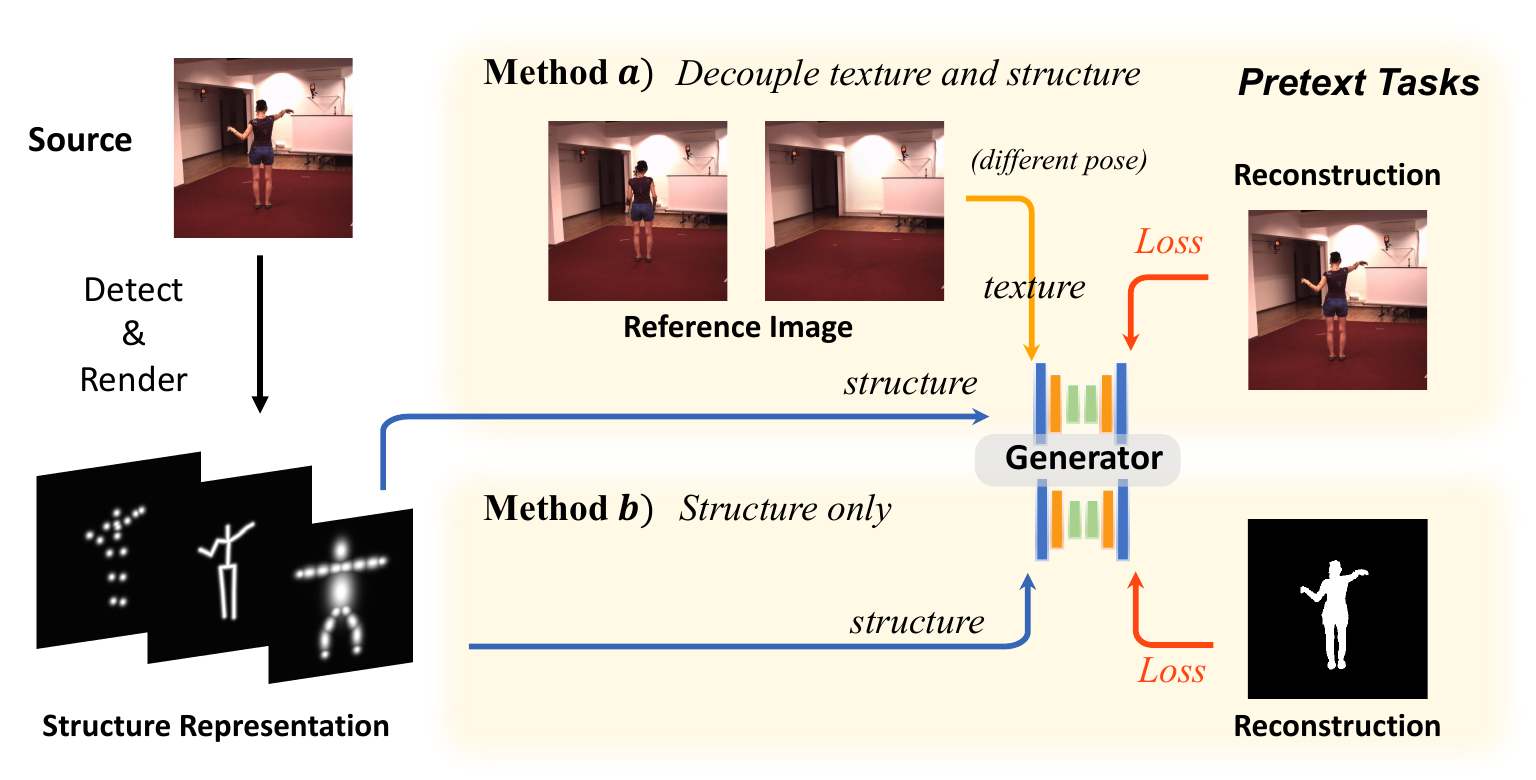}
        \caption{Illustration of prevailing pretext tasks in unsupervised monocular 3D pose estimation.}
        \label{fig:pretext-task}
    \end{subfigure}
    \begin{subfigure}[b]{0.5\linewidth}
        \centering
        \includegraphics[width=\textwidth]{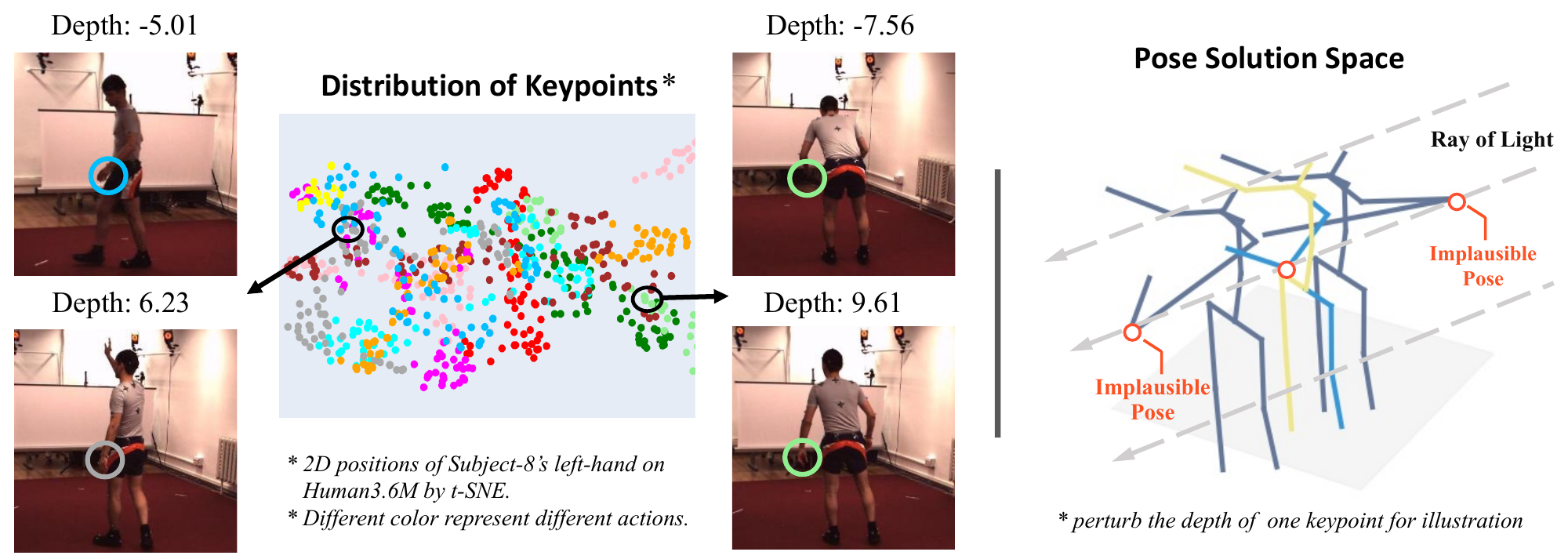}
        \caption{Illustration of the depth ambiguity issue. Left: depth visualization of neighbored embeddings from t-SNE~\cite{van2008visualizing} clusters of a representative 2D keypoint. Right: plausible and implausible 3D poses with the same 2D pose.
        }
        \label{fig:depth-ambiguity}
    \end{subfigure}

    \caption{Challenges. (a) illustrates the pretext tasks of related methods, which localize keypoints by various structure representations for human reconstruction. However, they are limited in 2D space without direct depth supervision. (b) highlights the widely observed depth ambiguity issue, where similar 2D poses correspond to distinctive relative depths from the pelvis and not all 3D correspondences along the ray of light are plausible for human structure. It remains an unexplored constraint problem in unsupervised monocular 3D pose estimation.}
    \label{fig:teaser}
\end{figure*}

One successful unsupervised learning technique, designing various pretext tasks, enables the detector to learn meaningful representation from a surrogate objective rather than paired ground truth.
In 3D pose estimation, an effective framework must satisfy two key requirements: accurate keypoint localization and proper human structure construction.
For keypoint localization, human image reconstruction (illustrated in~\cref{fig:pretext-task}) as a feasible pretext task involves training a generator that warps textures from a reference image or mask to the structure representations from keypoints modeled as Gaussian kernels~\cite{honari2022temporal,honari2022unsupervised} or bones as differentiable lines~\cite{he2022autolink}. 
This pretext task is effective because the generator needs sufficient spatial information from the structure representations to accurately reconstruct human images, which constrains the predicted keypoints located on the human body. 
In addition to spatial location, human structure demands unique body part correspondences for each keypoint and the body part order consistently aligning with a pre-given one. Physical priors are incorporated, including manually designed templates~\cite{kundu2020self,schmidtke2021unsupervised} and the connectivity of bones~\cite{yang2023mask} to constrain the keypoints.
With localization and structure consideration, previous methods have yielded promising results to some extent.

Nonetheless, these pretext tasks are designed in 2D space without direct supervision on depth dimension, ignoring the fact that monocular 3D pose estimation remains an inherently ill-posed problem due to depth ambiguity in 2D-to-3D.
The notorious ambiguity arises from the loss of depth information during the 3D-to-2D projection. For the inverse process, the same 2D point relates to multiple 3D points along the ray of light, which leads to a multi-solution nature. As illustrated in \cref{fig:depth-ambiguity}, similarly, neighboring 2D keypoints also have distinctive corresponding depths for monocular 3D pose estimation. Meanwhile, not all potential depths conforming to projective geometry are plausible due to the human skeleton structure.
Therefore, it is essential to develop a detector that can accommodate multiple solutions, while simultaneously adopting meticulous pretext tasks to impose depth constraints to narrow the pose solution space.

To address the aforementioned challenges, we propose a multi-solution tolerant detector and a series of novel pretext tasks within an integrated framework to contend with the depth ambiguity, named X as Supervision.
Motivated by the Multiple Choice Learning theory~\cite{NIPS2012multiple}, the detector handles the multi-solution problem by decoding multiple hypotheses within a local window from a single heatmap and leverages Winner-Takes-All loss~\cite{lee2016stochastic} to preserve the inherent multi-solution nature, while preventing the suppression of negative yet potentially correct samples. 
To apply depth constraints, we propose pretext tasks that leverage the parametric Skinned Multi-Person Linear (SMPL) model ~\cite{loper2023smpl} to generate plausible human structures.
The sampled data, formulated in a structured space informed by SMPL, is employed to regularize the distribution of predictions through a dual-representation discriminator based on Graph Convolution Network (GCN)~\cite{kipf2016semi}. Additionally, we render SMPL images as a complement to further constrain the detector. 

Extensive experiments conducted on human datasets Human3.6M~\cite{ionescu2013human3} and MPI-INF-3DHP~\cite{mono-3dhp2017} reveal that our approach surpasses existing state-of-the-art unsupervised 3D pose estimation methods. Further experiments highlight the ability to leverage in-the-wild data, showcasing the scalability of our approach. Additional experiments on animal datasets~\cite{biggs2020wldo} prove our approach's expansibility.

Our contributions are summarized in three-fold:

\begin{itemize}
    \item We emphasize the inherent depth ambiguity problem in the unsupervised monocular 3D pose estimation scenario. For the first time, we address the problem as a multi-solution issue via a novel multi-hypothesis detector.
    \item We introduce novel pretext tasks from SMPL-driven constraints with Winner-Takes-All loss to alleviate depth ambiguity. It fills the vacancy of direct 3D supervision in the previous pretext tasks.
    \item Our framework reveals remarkable robustness and generalization capability. Extensive experiments validate its effectiveness, achieving state-of-the-art performance on human datasets and promising results on animal datasets.
\end{itemize}
\begin{figure*}[th]
    \centering
    \includegraphics[width=0.95\linewidth]{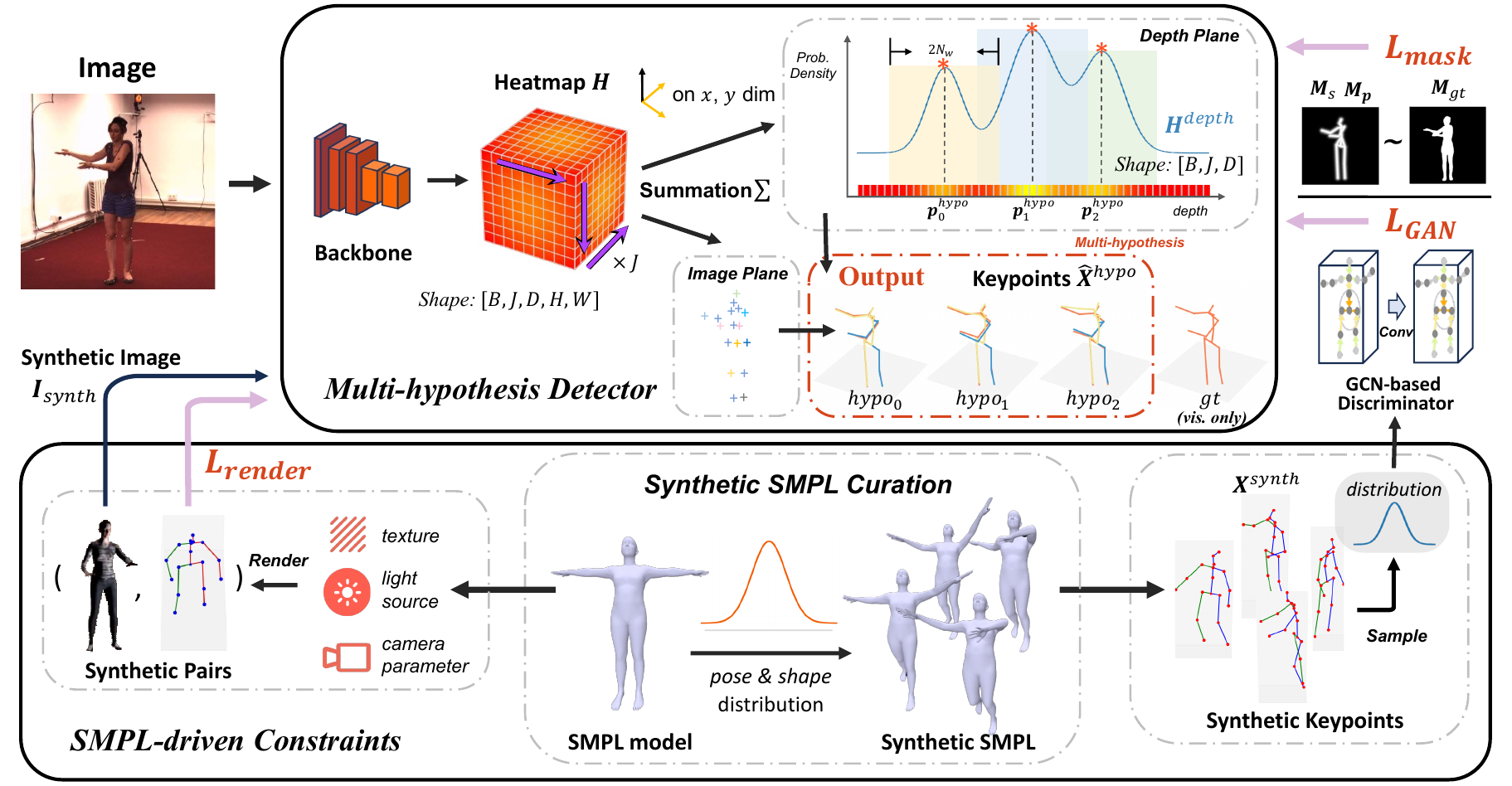}
    \caption{Framework overview of the X as Supervision. It contains a multi-hypothesis detector and novel pretext tasks for unsupervised training. The proposed detector takes in monocular images and novelly decodes multiple depth solutions by aggregating the local responses of the heatmap. For pretext tasks, in the 2D space, we adopt Masks as Supervision~\cite{yang2023mask} for keypoints localization and human structure. In the 3D space, we introduce human priors as constraints from a curated SMPL set via discriminative learning and synthetic pairs.}
    \label{fig:framework}
\end{figure*}

\section{Related Work}
\paragraph{Unsupervised Monocular 3D Pose Estimation.}
\label{sec:related-work}
Learning 3D poses from single-view images without manual annotations is the core objective of this task.
We limit the discussion scope and do not cover unsupervised 2D pose estimation~\cite{he2022autolink,jakab2018unsupervised} and lifting (2D pose given)~\cite{wandt2021canonpose,yu2021towards,srivastav2024selfpose3d}.

Unsupervised 3D pose estimation relies on pretext tasks to indirectly supervise the pose detector.
The primary pretext task, illustrated in \cref{fig:pretext-task}, involves rendering a structural representation from the detected pose, which is then used to reconstruct the human figure. This reconstruction process corrects the model's estimations in 2D.

However, achieving ``fully" unsupervised and monocular remains challenging, and several commonly adopted settings ease task difficulty but distract from the core concept:

\textbf{1) Reference Image/Multi-View (RI/MV)}. RI~\cite{kundu2020kinematic,kundu2020self} corresponds to Method a) in \cref{fig:pretext-task} which leverages reference images with the same subject to warp texture information for human reconstruction. 
MV~\cite{suwajanakorn2018discovery,sun2023bkind,honari2022unsupervised,yang2023mask} utilizes Triangulation~\cite{hartley2003multiple} in multi-view during training and supervises a monocular model, while overlooks depth ambiguity by using parallax.
It is limited to additional paired images for in-the-wild scenes. 

\textbf{2) Supervised Post-Processing (SPP)}.
SPP involves an ambiguous structure that renders keypoints as Gaussian kernel~\cite{sun2023bkind, honari2022unsupervised,honari2022temporal}. Without interrelationship consideration, permuting keypoint order will not yield changes in the loss value. It will predict high-dimensional landmarks (times larger than the ground truth) and train networks to map predictions to the ground truth. As raised in~\cite{he2022autolink,schmidtke2021unsupervised,yang2023mask}, it is limited to the heavy reliance on training labels.  

\textbf{3) Template (T)}. Kundu \etal \cite{kundu2020self} manually designed a template representation with various Gaussian kernels enforcing keypoint interpretability. It is labor-intensive and does not generalize well across diverse human physiques.

Apart from the common pretext task, Kundu \etal \cite{kundu2020kinematic} introduce a 3D degree of freedom (DoF) constraint at the bone level. Sosa \etal \cite{sosa2023self} constraint depth by 2D consistency of in-domain distribution from \textbf{4) unpaired joints (J)} after 3D random rotation and projection. However, these indirect constraints do not explicitly involve depth information which may risk mode collapse in training.

In this work, we advance unsupervised monocular 3D pose estimation by designing direct depth constraints, avoiding reliance on additional view and manual annotations in the above settings.
\section{Method}
\subsection{Monocular 3D Pose Estimation}
For clarity, we initiate the discussion with the general formulation of the monocular 3D pose estimation task.
Given an input image $\mathbf{I}$, the goal is to estimate $J$ number of joint locations $\hat{\mathbf{X}} \in \mathbb{R}^{J \times 3}$ in the 3D coordinate system:
\begin{equation}
    \hat{\mathbf{X}} = \phi(\mathbf{I}),
\end{equation}
where network $\phi$ is optimized to approximate ground truth $\mathbf{X}^{gt}$ during training.

Vanilla heatmap-based detector~\cite{sun2018integral} typically predicts a 3D heatmap $\mathbf{H} \in \mathbb{R} ^{J \times D \times H \times W}$, where $D$, $H$, and $W$ indicate the depth, height, and width dimensions, respectively, to represent the probability of joint locations. The 3D joint locations $\hat{\mathbf{X}} $ are obtained by maximum likelihood in a soft aggregation:
\begin{equation}
\label{eq:vanilla-detector}
    \hat{\mathbf{X}} = \sum^D_{p_z=1}\sum^H_{p_y=1}\sum^W_{p_x=1} \mathbf{p} \cdot \sigma(\mathbf{H}(\mathbf{p})),
\end{equation}
where $\mathbf{p}$ indicates the position index of the heatmap and $\sigma$ is the Softmax function. 

\subsection{Multi-hypothesis Detector}
\label{sec:multi-hypo-det}
Due to the depth ambiguity in unsupervised monocular 3D pose estimation, multiple plausible depth solutions from samples with similar 2D pose predictions, activate different regions of the heatmap during training. 
Therefore, the ideal heatmap that contains various potential depth estimations should naturally exhibit a multimodal distribution\footnote{\scriptsize{\url{https://en.wikipedia.org/wiki/Multimodal_distribution}}}, characterized by multiple peaks in response values. 
However, the vanilla detector enforces predictions to a single condition, potentially suppressing these critical positive responses.

To address this, as demonstrated in the top part of \cref{fig:framework}, we formulate the task as a multi-solution problem that each peak located on the heatmap implies a potential estimate and propose a multi-hypothesis detector that outputs $N_{hypo}$ of joint locations $\hat{\mathbf{X}}^{hypo} \in \mathbb{R}^{N_{hypo} \times J \times 3}$.

Specifically, we leverage the probability heatmap on the $z$ dimension, aiming to decode hypotheses from depth distribution.
The probability depth heatmap, denoted as $\mathbf{H}^{depth} \in \mathbb{R} ^ {J \times D}$ is derived from the origin heatmap $\mathbf{H}$ via summation:
\begin{equation}
    \mathbf{H}^{depth} = \sum^H_{p_y=1}\sum^W_{p_x=1} \sigma(\mathbf{H}(\mathbf{p})).
\end{equation}

From this, candidate peaks are decoded and then selected based on their response values. The candidate peaks, $\mathbf{p}^{cand}$, are identified using the condition: 
\begin{equation}
\label{eq:peak-selection}
    \mathbf{p}^{cand}_{i} = \mathbf{1}(\mathbf{H}^{depth}_i \geq \mathbf{H}^{depth}_{i-1} \land \mathbf{H}^{depth}_i \geq \mathbf{H}^{depth}_{i+1}),
\end{equation}
where the index $i$ ranges in $[1, D-2]$ and $\mathbf{1}$ indicates a logical operation. Subsequently, $N_{hypo}$ of the hypothesis peaks $\mathbf{p}^{hypo}$ are determined by selecting the top-$K$ candidates:
\begin{equation}
    \mathbf{p}^{hypo} = topK(\mathbf{H}^{depth}(\mathbf{p}^{cand}), N_{hypo}).
\end{equation}

Directly taking the peak locations as the prediction of joint locations may result in quantization error due to the heatmap resolution being much lower than the original image resolution. Correspondingly, we implement local weighting to refine the locations. With a window size of $2 N_{w}$, multiple depth hypotheses $\hat{\mathbf{z}}^{hypo}$ of joint location $\hat{\mathbf{X}}^{hypo}$ are computed as:
\begin{equation}
\label{eq:depth-hypotheses-weighting}
    \hat{\mathbf{z}}^{hypo}_h = {\mathop{\sum}^{i=w_h^s}_{i=w_h^e} \mathbf{p}_i \cdot \mathbf{H}^{depth}(\mathbf{p}_i)} / {\mathop{\sum}^{i=w_h^s}_{i=w_h^e} \mathbf{H}^{depth}(\mathbf{p}_i)},
\end{equation}
where $w_h$ denotes the local window for the $h^{th}$ hypothesis, with its start and end index $\{\mathbf{p}^{hypo}_h - N_{w}, \mathbf{p}^{hypo}_h + N_{w}\}$. $x$ and $y$ dimensions are preserved as the original in \cref{eq:vanilla-detector}.

To match the multi-solution characteristics of the task,
we adopt Winner-Takes-All (WTA) loss~\cite{lee2016stochastic}, $\mathcal{L}^{WTA}$ on the depth dimension for our proposed pretext tasks:
\begin{equation}
    \mathcal{L}^{WTA} = \mathop{\min}_{h\in N_{hypo}}(\mathcal{L}(\hat{\mathbf{X}}^{hypo}_h)),
\end{equation}
which selectively penalizes a single local response during each backpropagation and maintains the multi-modal characteristic of distribution.

Unlike multi-hypothesis methods in supervised learning~\cite{li2021human,li2022mhformer}, where multiple hypotheses are treated as intermediate results, our proposed detector keeps them as output for the inherent multi-solution property.
In contrast to the multi-head design, our approach only modifies the decoding process on the heatmap which can be consolidated into matrix operations (see \cref{sec:supp-matrix-conversion}), incurring no additional overhead on the network.

\begin{figure}
    \centering
    \includegraphics[width=0.95\linewidth]{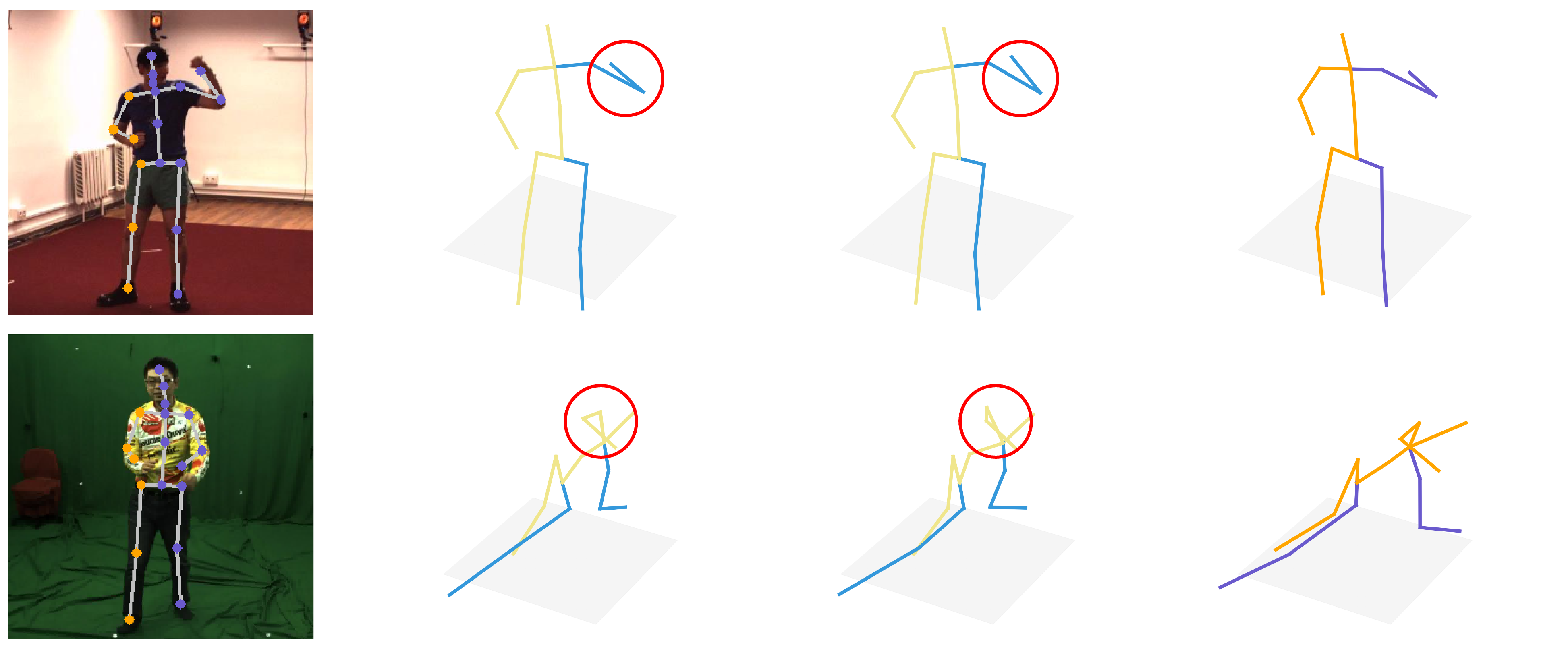}
    \caption{Visualization of the predicted multiple hypotheses on Human3.6M~\cite{ionescu2013human3} and MPI-INF-3DHP~\cite{mono-3dhp2017}. Differences are highlighted in red circles. The ground truth poses are plotted in the last column. More visualization results are presented in \cref{sec:supp-qualitative}.}
    \label{fig:mh}
\end{figure}

\subsection{Unsupervised Pretext Tasks}
Up to this point, the proposed multi-hypothesis detector has been discussed as a foundation to handle depth ambiguity.
As shown in \cref{fig:mh}, multi-hypothesis with appropriate constraints effectively captures subtle details, resulting in improved performance in evaluation.
Next, as illustrated in the bottom part of \cref{fig:framework}, we will demonstrate how the constraints of human priors can be introduced for supervision. 

\subsubsection{Preliminaries}
\paragraph{Mask as Supervision Revisited.}
\label{sec:mask-as-sup}
Easily attainable human mask $\mathbf{M}_{gt}$ is leveraged for pretext supervision to localize 2D keypoints in \cite{yang2023mask}. As illustrated in ~\cref{fig:pretext-task} Method b), it renders a Skeleton Mask $\mathbf{M}_s$ from predicted keypoints $\hat{\mathbf{x}}$ on the 2D image plane and then generates a Physique Mask $\mathbf{M}_p$ with a reconstruction network.
The network is optimized with the objective function:
\begin{equation}
    L_{mask} = \lambda_s ||\mathbf{M}_{gt} - \mathbf{M}_s||_2^2 + \lambda_p ||\mathbf{M}_{gt} - \mathbf{M}_p||_2^2,
\label{eq:mask}
\end{equation}
where $\lambda_s$ and $\lambda_p$ are balancing hyper-parameters.

To approximate the ground truth mask $\mathbf{M}_{gt}$, Skeleton Mask $\mathbf{M}_s$ should stretch through the foreground area for Physique Mask $\mathbf{M}_p$ reconstruction.
This necessitates the accurate location of the predicted keypoints, which serve as the rendering source for the Skeleton Mask.

From a top-down design of estimation, we employ this effective approach as the pretext task on the 2D plane.
Given our primary focus on resolving depth ambiguity, the WTA loss design is not applied to this 2D loss.

\vspace{-3mm}
\paragraph{SMPL Revisited.} Digitalizing a compatible 3D human body with high fidelity is a popular topic in both computer graphics and computer vision. In this domain, one of the most widely used avatar models, the Skinned Multi-Person Linear (SMPL) model~\cite{loper2023smpl} systematically captures human body deformation by 72-dimensional pose parameter 
$\boldsymbol{\theta}$  
defined to represent 23 joints and root rotation through Rodrigues form, and 10-dimensional shape parameter
$\boldsymbol{\beta}$ 
characterized in normalized unit vectors via Principal Component Analysis (PCA). 3D locations of human mesh vertices $V\in \mathbb{R}^{6890\times 3}  $ are computed as:
\begin{equation}
\label{eq:smpl}
    V = \rm{SMPL}(\boldsymbol{\theta}, \boldsymbol{\beta}, \xi),
\end{equation}
where $\xi$ indicates the pose and shape blend parameters.

SMPL offers the advantage of a parametric form with realistic and drivable mesh. It encapsulates 3D human priors, reducing the complex human pose locations in unbound 3D space into a manageable low-dimensional control space.

\subsubsection{SMPL-driven Constraints}
In unsupervised monocular 3D pose estimation, constraints in 3D are necessary due to depth ambiguity but hard to acquire from 2D images.
Meanwhile, the structured parameter space of SMPL enables obtaining 3D information via a constructed distribution of parameters.

Inspired by the features of SMPL, we introduce SMPL-driven constraints in an unsupervised fashion to capture these 3D priors. 
It comprises two primary components: curation of SMPL data to build a probability distribution for plausible and diverse poses and employment of sampled 3D data as pretext tasks for supervisory.

\vspace{-3mm}
\paragraph{Synthetic SMPL Curation.}
Building on the capabilities of SMPL, the key focus of curation is constructing a plausible data distribution.
Recognizing that human poses and body shapes vary in frequency, with certain poses occurring more often and accounting for the majority,
the Gaussian distribution is well-suited to capture a range of variations around a mean value.
Particularly, we model each SMPL parameter as an independent truncated Gaussian distribution, where the random parameter conformed to Gaussian distribution with the mean $\mu$ and variance $\sigma$ is constrained on an interval $(\gamma_l, \gamma_u)$.

From this, a set of SMPL mesh is generated according to \cref{eq:smpl}. With the regression matrix $\mathcal{J}$ provided by SMPL, the sampled data is converted to the desired 3D keypoints $\mathbf{X}^{synth}$:
\begin{equation}
    \mathbf{X}^{synth} = \mathcal{J}(V,I),
\end{equation}
where $V$ and $I$ are vertices and indices of SMPL mesh.

With the same target to ease data scarcity, several synthetic human datasets~\cite{varol2017learning,patel2021agora,yang2023synbody,ge20243d} have been proposed to boost the performance of model~\cite{STRAPS2020BMVC,purkrabek2024improving}. These datasets serve as baselines to assess the effectiveness of data curation for 3D pose estimation.

\vspace{-3mm}
\paragraph{Constraints Employment.}
To introduce 3D prior knowledge from the SMPL set into the detector, we use adversarial learning to impose human structure constraints based on 3D keypoints $\mathbf{X}^{synth}$.
Leveraging the graph's effectiveness in topology modeling \cite{shi2019two, shi2019skeleton}, we employ a GCN-based discriminator $\psi$ to regularize the distribution of the detector's predictions, emphasizing the interrelationships of the human skeletal structure. The discriminator is composed of stacked GraphSAGE~\cite{hamilton2017inductive} layers (refer to \cref{sec:supp-gcn} for the details of network structure).  To enhance representation, we design two undirected graphs $G_k = (V_k, E_k)$ and $G_b = (V_b, E_b)$ to embed keypoint-level and bone-level input. For keypoints $\mathbf{X}$, 3D locations are set as nodes, and the edges are defined following human skeleton connectivity $S$. We keep $S$ the same as the one utilized in \cref{sec:mask-as-sup}.
For bones $\mathbf{B}$, nodes are computed as the difference of two connected keypoints. 
The dual-representation graph is constructed as: 
\begin{gather}
    V_k = \{X_i| i=1,...,J\}, \\
    V_b = \{X_i - X_j|(i,j) \in S\}, \\
    E_k = S, \ E_b = S.
\end{gather}

The objective function $L_{GAN}$ integrates the generator loss  $L^{WTA}_\phi$ and discriminator loss $L^{WTA}_\psi$, following LS-GAN~\cite{mao2017least}  with a Winner-Takes-All approach. Here, the detector $\phi$  acts as the generator in the GAN setting.
\begin{equation}
\begin{aligned}
    L^{WTA}_\psi = & \ \frac{1}{2} \mathbb{E} \left[(\psi([\mathbf{X}^{synth}, \mathbf{B}^{synth}]) - 1)^2\right] \\ & + \frac{1}{2} \mathbb{E} \left[\psi([\hat{\mathbf{X}}^{hypo}, \hat{\mathbf{B}}^{hypo}])^2\right],   
\end{aligned}
\label{eq:gan}
\end{equation}
\begin{equation}
\begin{aligned}
    L^{WTA}_\phi = \frac{1}{2} \mathbb{E} \left[(\psi([\hat{\mathbf{X}}^{hypo}, \hat{\mathbf{B}}^{hypo}]) - 1)^2\right].
\end{aligned}
\end{equation}
The GAN-based loss with WTA provides a relatively loose constraint on the 3D structure of prediction. It only enforces one of the hypotheses close to the sampled case where multiple potential depth conditions are included by sampling.

Furthermore, SMPL mesh can also be integrated with the light source, human texture, and camera parameters, allowing SMPL mesh to be rendered as a synthetic human image $\mathbf{I}_{synth}$ with corresponding 3D keypoint locations. A direct $\mathcal{L}_2$ loss for regression is employed:
\begin{equation}
    L^{WTA}_{render} = ||\mathbf{X}^{synth} - \phi(\mathbf{I}_{synth})||_2^2.
    \label{eq:render}
\end{equation}
Based on the curated distribution of SMPL parameters, the divergence of paired data will be further enlarged by different light and random camera parameters~\cite{pavlakos2018learning}. It helps the model to understand 2D-to-3D on diverse appearances and perspectives and also benefits stabilizing GAN training.

\begin{figure*}
\begin{minipage}{\textwidth}
    \begin{minipage}{0.68\textwidth}
    {\small
        \captionof{table}{Comparison with state-of-the-art methods on Human3.6M. As discussed in~\cref{sec:related-work}: \textbf{SPP}: supervised post-processing. \textbf{RI/MV}: reference image or multi-view. \textbf{T}: template. \textbf{J}: unpaired ground truth joints. Our baselines are introduced in~\cref{sec:our-baseline}. MPJPEs are in $mm$.}
        \begin{tabular}{l|c|c|c|c|c|c|c}
        \toprule
        \multirow{2}{*}{Method} & \multicolumn{4}{|c|}{\textbf{Settings}} & \multicolumn{3}{c}{\textbf{Metrics ($\bm{\downarrow}$)}}\\
        \cmidrule{2-8}
        {} &  {RI/MV} & {SPP} & {T} & {J} & {MPJPE} & {N-MPJPE} & {P-MPJPE} \\
        \midrule

        {Suwajanakorn \etal~\cite{suwajanakorn2018discovery}} & {\color{red}{\checkmark}} & {\color{red}{\checkmark}} & {\color{green}{$\times$}} & {\color{green}{$\times$}} & {$158.7$} & {$156.8$} & {$112.9$} \\

        {Sun \etal~\cite{sun2023bkind}} & 
        {\color{red}{\checkmark}} & {\color{red}{\checkmark}} & {\color{green}{$\times$}} & {\color{green}{$\times$}} & {$125.0$} & {-} & {$105.0$} \\

        {Honari \etal~\cite{honari2022temporal}} &
        {\color{green}{$\times$}} & {\color{red}{\checkmark}} & {\color{green}{$\times$}} & {\color{red}{\checkmark}} & {$100.3$} & {$99.3$} & {$74.9$} \\
    
        {Honari \etal~\cite{honari2022unsupervised}} &
        {\color{red}{\checkmark}} & {\color{red}{\checkmark}} & {\color{green}{$\times$}} & {\color{green}{$\times$}} & {$73.8$} & {$72.6$} & {$63.0$} \\
    
        \midrule
        {Sosa \etal~\cite{sosa2023self}} & {\color{green}{$\times$}} & {\color{green}{$\times$}} & {\color{green}{$\times$}} & {\color{red}{\checkmark}} & {-} & {-} & {$96.4$} \\
    
        {Kundu \etal~\cite{kundu2020self}}  & {\color{red}{\checkmark}} & {\color{green}{$\times$}} & {\color{red}{\checkmark}} & {\color{red}{\checkmark}} & {$99.2$} & {-} & {-} \\
    
        {Kundu \etal~\cite{kundu2020kinematic}}  & {\color{red}{\checkmark}} & {\color{green}{$\times$}} & {\color{green}{$\times$}} & {\color{green}{$\times$}}& {-} & {-} & {$89.4$} \\

        {Yang \etal~\cite{yang2023mask}} & {\color{red}{\checkmark}} & {\color{green}{$\times$}} & {\color{green}{$\times$}} & {\color{green}{$\times$}} & {$95.9$} & {$96.8$} & {$90.4$} \\

        \midrule
        {Ours-SynSH} & {\color{green}{$\times$}} & {\color{green}{$\times$}} & {\color{green}{$\times$}} & {\color{green}{$\times$}} & {$93.0$} & {$92.7$} & {$72.2$} \\ 
        {Ours-SynMH (conf)} & {\color{green}{$\times$}} & {\color{green}{$\times$}} & {\color{green}{$\times$}} & {\color{green}{$\times$}} & {$92.6$} & {$92.4$} & {$75.2$} \\ 
        {Ours-SynMH (best)} & {\color{green}{$\times$}} & {\color{green}{$\times$}} & {\color{green}{$\times$}} & {\color{green}{$\times$}} & {\textcolor{gray}{$76.9$}} & {\textcolor{gray}{$76.6$}} & {\textcolor{gray}{$67.0$}} \\ 

        \midrule
        {Ours-SurSH} & {\color{green}{$\times$}} & {\color{green}{$\times$}} & {\color{green}{$\times$}} & {\color{green}{$\times$}} & {$94.1$} & {$93.7$} & {$62.8$} \\ 
        {Ours-SurMH (conf)} & {\color{green}{$\times$}} & {\color{green}{$\times$}} & {\color{green}{$\times$}} & {\color{green}{$\times$}} & {$93.7$} & {$93.4$} & {$60.9$} \\ 
        {Ours-SurMH (best)} & {\color{green}{$\times$}} & {\color{green}{$\times$}} & {\color{green}{$\times$}} & {\color{green}{$\times$}} & {\textcolor{gray}{$\mathbf{72.2}$}} & {\textcolor{gray}{$\mathbf{71.8}$}} & {\textcolor{gray}{$\mathbf{56.1}$}} \\
        \bottomrule
      \end{tabular}
      \label{table:hm36-table}
    }
    \end{minipage}
    \hfill
    \begin{minipage}{0.3\textwidth}

        \centering
        \includegraphics[width=0.8\linewidth]{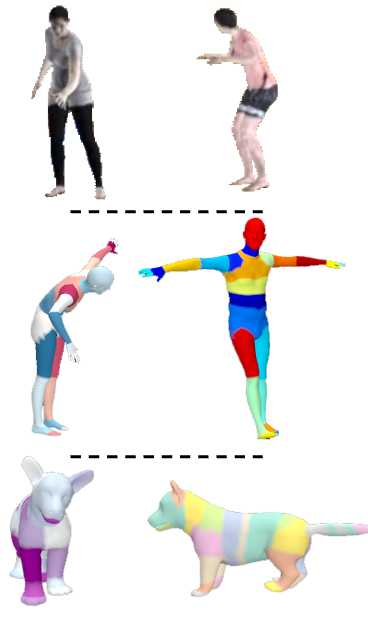}
        \caption{Illustration of different textures. From top to down: SURREAL~\cite{varol2017learning} human texture, part segmentation of SMPL~\cite{loper2023smpl} and SMAL~\cite{Zuffi2017smal}.}
        \label{fig:texture}

    \end{minipage}
\end{minipage}
\end{figure*}

\subsubsection{Loss}
According to \cref{eq:mask}, \cref{eq:gan} and \cref{eq:render}, the final objective for the detector is defined as a weighted combination of the previous losses:
\begin{equation}
    \mathcal{L} = \lambda_m L_{mask} + \lambda_g L^{WTA}_\phi + \lambda_r L^{WTA}_{render},
\end{equation}
where $\lambda_m$, $\lambda_g$ and $\lambda_r$ are hyper-parameters to balance training losses.

The final objective organizes 2D keypoints structure and localization by \cref{eq:mask} on the training data while offering loose 3D constraints from \cref{eq:gan} and \cref{eq:render}, enabling effective unsupervised learning with depth ambiguity.

\section{Experiments}
\subsection{Datasets}
We use the following datasets for training and evaluation. 
Note that, our approach strictly follows the unsupervised setting that all ground truth poses are not accessible.
Implementation details can be found in the \cref{sec:spp-implemenation}.

\noindent\textbf{Human3.6M}~\cite{ionescu2013human3}.
Following~\cite{zhang2018unsupervised}, six activities (direction, discussion, posing, waiting, greeting, walking) are selected, and we train the model on subjects 1, 5, 6, 7, 8, evaluate on subjects 9, and 11.

\noindent\textbf{MPI-INF-3DHP}~\cite{mono-3dhp2017}.
Following \cite{honari2022unsupervised}, subjects 1 to 6 are for training, and 7 and 8 are for evaluation. Frames are filtered out where the person is occluded~\cite{yang2023mask}.

\noindent\textbf{TikTok}~\cite{jafarian2021learning} and \textbf{MPII}~\cite{andriluka14cvpr}.
TikTok comprises single-view video sequences from social media with large appearance diversity, without any pose annotations. MPII collects in-the-wild single-view images with 2D pose annotation. We follow~\cite{yang2023mask} to process these datasets.

\noindent\textbf{StandfordExtra}~\cite{biggs2020wldo}. StanfordExtra provides 2D keypoint and segmentation annotations for 12,000 images of 120 dog breeds. We utilize it to verify the ability of the proposed unsupervised learning approach on non-human species. 

\subsection{Comparison with State-of-the-Art Methods}
\subsubsection{Baselines}
\label{sec:our-baseline}
To illustrate the performance of our proposals, we comprehensively organize several baselines and abbreviate them as follows.
For the detector,
\textit{SH} indicates the determinate baseline of the vanilla single-hypothesis detector. \textit{MH} indicates the proposed multi-hypothesis detector, where \textit{conf} takes the hypothesis with the maximum peak response and \textit{best} utilizes the hypothesis closest to the ground truth, performing the upper bound for addressing depth ambiguity. 
For the SMPL-driven constraints, we also compare with different synthetic settings:
\textit{Syn} indicates the proposed distribution of SMPL parameters, and \textit{Sur} indicates the existing synthetic SMPL-based dataset, SURREAL~\cite{varol2017learning}.

\subsubsection{Results on Human3.6M}
In \cref{table:hm36-table}, we present and clarify experimental settings (RI/MV, SPP, T, and J) of the prevailing methods. As introduced in \cref{sec:related-work}, they deviate from unsupervised monocular 3D pose estimation by involving additional information. Instead, we take a step forward that does not utilize any of these techniques and obeys the fundamental setting.

In \cref{table:hm36-table}, several experimental observations can be concluded as: 

\noindent
\textbf{Multi-hypothesis analysis.} 
1) Without leveraging the previously mentioned tricks, our \textit{SH} baselines achieve competitive results. The corresponding \textit{MH} model improves the performance by a large margin. 
2) The \textit{MH} model can capture the 3D structure better by maintaining the inherent ambiguity nature. Even the most confident one in multiple hypotheses (\textit{conf}) outperforms the single-hypothesis one (\textit{SH}). Meanwhile, as \textit{best} performs better than \textit{conf}, we can conclude that more plausible potential depths are discovered in multiple hypotheses.

\noindent
\textbf{Pretext task analysis.} The proposed SMPL-driven constraints effectively supervise the detector in both the curated distribution (\textit{Syn}) and a carefully obtained one (\textit{Sur}).

\begin{table}[t]
    \centering
 {\small
    \captionof{table}{Comparison with state-of-the-art methods on MPI-INF-3DHP. MPJPE is in $cm$. Note that the first four methods utilize SPP, the first five methods involve RI/MV, and Sosa \etal \cite{sosa2023self} use J for their results with additional Human3.6M data, while we keep the setting under monocular and unsupervised.}
    \begin{tabular}{l|c|c|c}
    \toprule
    {Method} & {PCK($\uparrow$)} & {AUC($\uparrow$)} & {MPJPE($\downarrow$)} \\
    \midrule
    {Denton \etal \cite{denton2017unsupervised}} & {-} & {-} & {$22.28$} \\
    {Rhodin \etal \cite{rhodin2019neural}} & {-} & {-} & {$20.24$} \\
    {Honari \etal \cite{honari2022temporal}} & {-} & {-} & {$20.95$} \\
    {Honari \etal \cite{honari2022unsupervised}} & {-} & {-} & {$14.57$} \\
    \midrule
    {Yang \etal \cite{yang2023mask}} & {$60.2$} & {$24.7$} & {$19.36$} \\
    \midrule
    {Sosa \etal \cite{sosa2023self}} & {$69.6$} & {$32.8$} & {-} \\
    \midrule
    {Ours-SynMH (conf)} & {$65.2$} & {$31.0$} & {$15.85$} \\
    {Ours-SynMH (best)} & {\textcolor{gray}{$71.7$}} & {\textcolor{gray}{$36.6$}} & {\textcolor{gray}{$13.48$}} \\
    {Ours-SurMH (conf)} & {$\mathbf{70.9}$} & {$\mathbf{35.6}$} & {$\mathbf{13.86}$} \\
    {Ours-SurMH (best)} & {\textcolor{gray}{$\mathbf{78.5}$}} & {\textcolor{gray}{$\mathbf{42.2}$}} & {\textcolor{gray}{$\mathbf{11.50}$}} \\
    \bottomrule
  \end{tabular}
  \label{table:mpi-inf-3dhp-table}
}
\end{table}

\subsubsection{Results on MPI-INF-3DHP}
Enabling more in-depth exploration in complex scenarios, we evaluate our proposed approach's effectiveness on MPI-INF-3DHP~\cite{mono-3dhp2017}, a more challenging dataset. In \cref{table:mpi-inf-3dhp-table}, it demonstrates consistent superiority when facing more diverse actions and appearances. Further qualitative results are illustrated in \cref{sec:supp-qualitative}.

\subsection{Ablation Study}
To thoroughly evaluate the proposed pretext task of SMPL-driven constraints, we conducted ablation studies based on the following core questions:

\noindent
\textbf{Q1}: How effective of the discriminator in $L^{WTA}_\phi$ for conveying SMPL priors to the detector?

\noindent
\textbf{Q2}: How do the pose and appearance affect the results, specifically the form of the curated set in $L^{WTA}_{render}$?

Note that, as the proposed unsupervised framework is integrated, instead of simply removing components step-by-step, we construct corresponding baselines to reflect the concepts of design and analyze each module. All the experiments are conducted on Human3.6M dataset.

\vspace{-3mm}
\paragraph{Ablation study on discriminative learning.} 
In our analysis \cref{table:disc-ablation}, we choose the vanilla detector (\textit{SH}) to avoid additional effects and establish juxtaposed discriminative strategies as baselines while keeping the rest of the pretext tasks unchanged. In detail, \ding{172} removes the discriminative pretext task. \ding{173} uses an MLP-based discriminator. \ding{174} takes 2D ground truth joints for direct supervision as an alternative, reflecting the previous 2D pretext tasks. \ding{175} follows our proposed method. 
Using Triangulation~\cite{hartley2003multiple}, we can obtain a relative accuracy for 2D localization, as the 3D results from Triangulation fully rely on the 2D.

Compared with \ding{172}, with effective 3D constraints, the model gains a significant performance improvement. 
Compared with \ding{173}, the proposed dual-representation GCN-based discriminator delivers a more robust representation.

Further investigation reveals the importance of 3D constraints in alleviating depth ambiguity. In \ding{172} and \ding{174}, replacing 3D discriminative loss with paired 2D ground truth loss yields higher Triangulation accuracy, but the improvement gap narrows in the monocular. It implies the supervision of 2D alone is insufficient due to the depth ambiguity and the proposed method successfully alleviates it by 3D constraint.

\begin{table}[t]
    \centering
 {\small
\captionof{table}{Ablation study on discriminative learning. $/$ indicates the results of Triangulation. Disc. indicates discriminative constraint. $w$ and $w/o$ indicate $with$ and $without$, respectively.}
    \begin{tabular}{l|c|c}
    \toprule
    {Baseline} & {MPJPE($\downarrow$)} & {Improvement}\\
    \midrule
    {\ding{172} $w/o$ Disc.} & {$121.16_{/48.42}$} & {-} \\
    {\ding{173} vanilla Disc.} & {$95.63_{/41.83}$} & {$+25.53_{/+6.59}$}\\
    {\ding{174} $w/o$ Disc.; $w$ 2D GT} & {$91.37_{/28.94}$} & {$+29.79_{/+19.48}$} \\
    \midrule
    {\ding{175} ours Disc.} & {$94.14_{/41.69}$} & {$+27.02_{/+6.73}$}\\
    \bottomrule
  \end{tabular}
  \label{table:disc-ablation}
}
\end{table}

\begin{table}[t]
    \centering
 {\small
    \captionof{table}{Ablation study on the quality of synthetic set.}
    \begin{tabular}{l|c}
    \toprule
    {Baseline} & {MPJPE($\downarrow$)} \\
    \midrule
    {Syn-Seg} & {$113.48$} \\
    {Syn-Texture} & {$105.61$} \\
    {Sur-Texture} & {$88.56$} \\
    \bottomrule
  \end{tabular}
  \label{table:synth-set-ablation}
}
\end{table}

\vspace{-3mm}
\paragraph{Ablation study on synthetic set.}
We select the best implementation, MH (best) to focus on verifying the synthetic set quality on rendering results.
Ablations are split into distribution settings (\textit{Syn} and \textit{Sur}), and appearance settings (\textit{Seg} and \textit{Texture}).
As illustrated in \cref{fig:texture}, we adopt random normalized color maps based on the part segmentation information provided in SMPL as appearance baselines, \textit{Seg}. \textit{Texture} indicates the original synthetic human appearance from UV maps. As expected, compared with the previous line for the last two rows of \cref{table:synth-set-ablation}, refined distribution (\textit{Sur}) and appearance (\textit{Texture}) lead to a better result.
Notably, with the core idea of the proposed pretext task in place, our method in the alternative (\textit{Syn-Seg}) setting still presents competitive performance, showcasing its effectiveness.
Experimentally, with the discriminative constraint, our method achieves a satisfying MPJPE of $107.9 mm$. It is subsequently utilized for extended experiments on animals when the detailed texture is not accessible.

\begin{figure}[t]
    \centering
    \includegraphics[width=\linewidth]{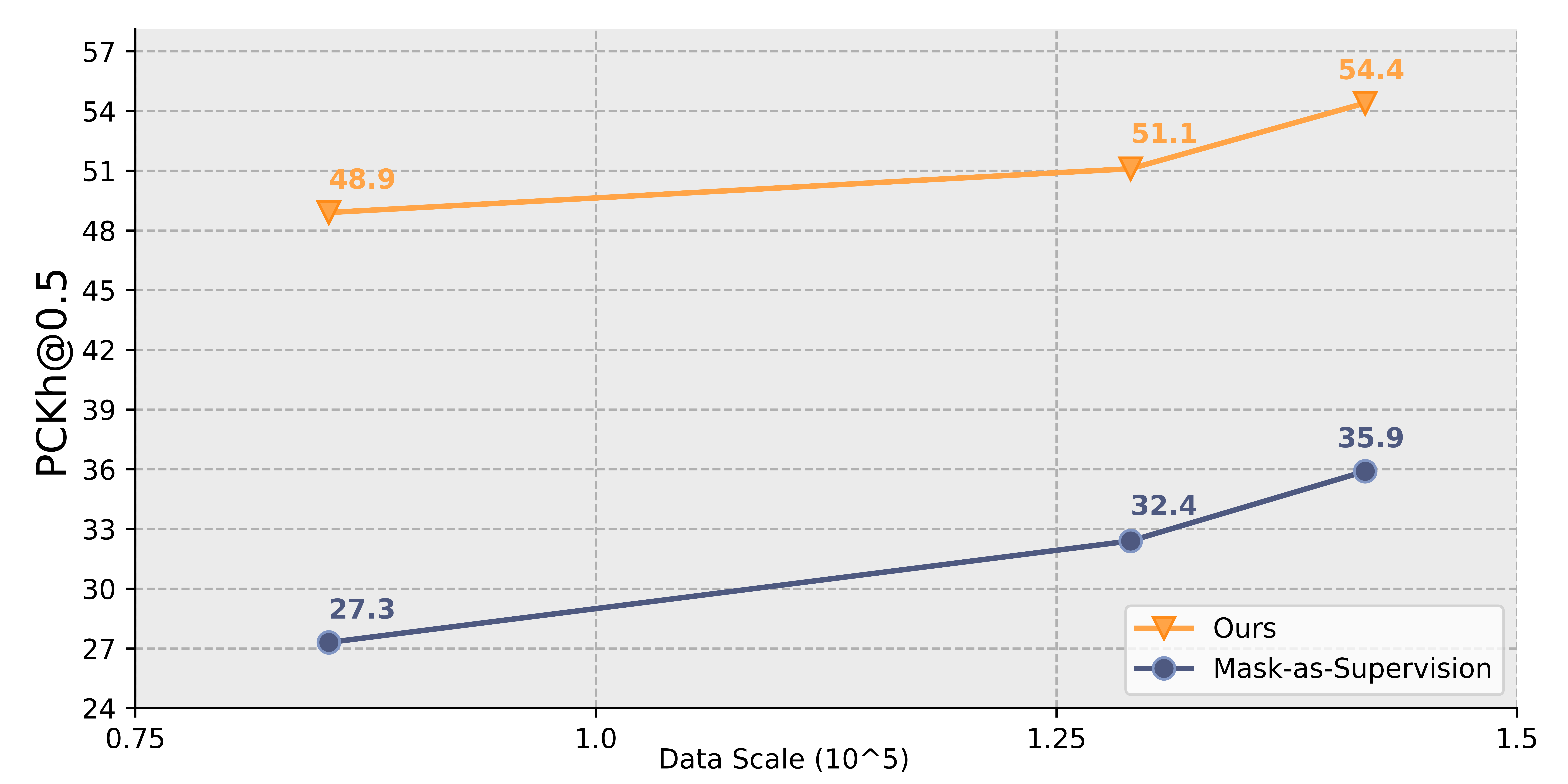}
    \caption{Performance on in-the-wild MPII dataset. Each node indicates that a dataset is progressively adopted for data scale-up training. MPII dataset is only utilized for evaluation.}
    \label{fig:scale}
\end{figure}

\begin{figure}[t]
    \centering
    \includegraphics[width=\linewidth]{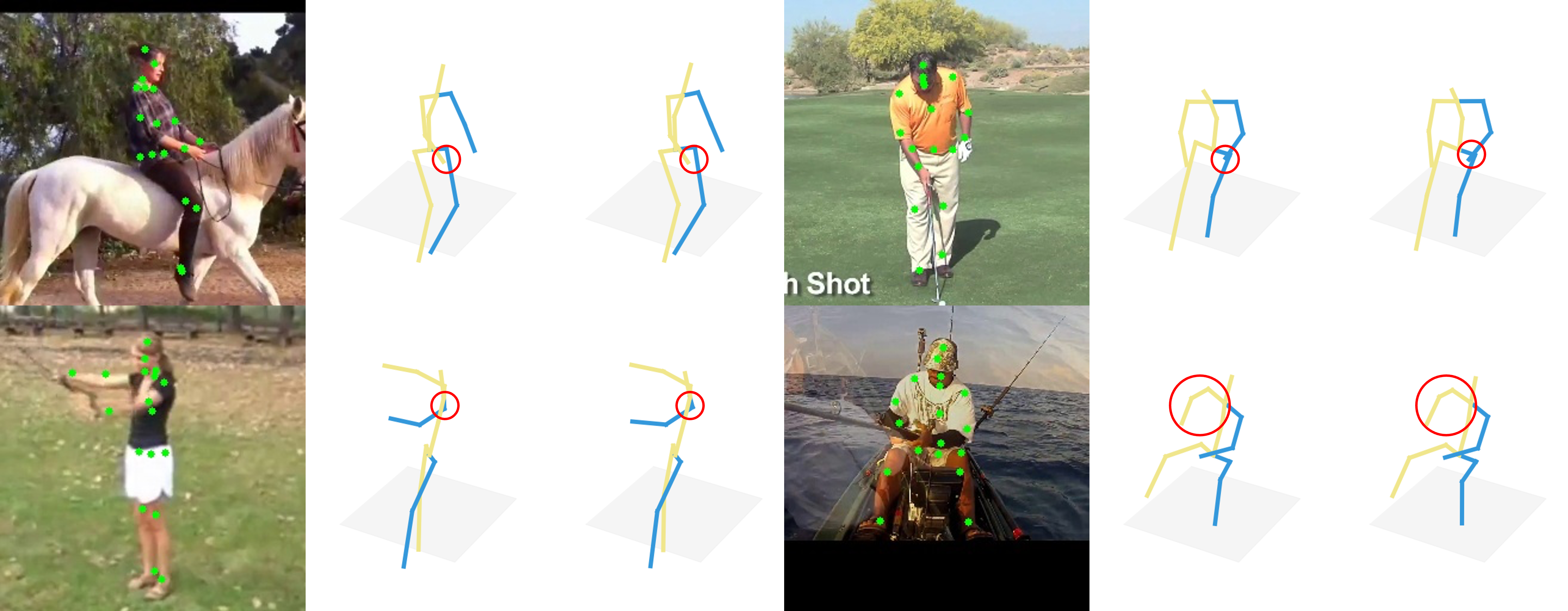}
    \caption{Visualization of the predicted 2D projected poses and 3D pose hypotheses on the MPII dataset. Top-2 hypotheses are selected. Differences are highlighted in red circles.}
    \label{fig:mpii}
\end{figure}

\subsection{Generalization Analysis}
To satisfy the core objective of unsupervised learning, a proposed method should be both scalable to handle large amounts of data and adaptable across diverse scenarios. We therefore conduct experiments focused on these key aspects.

\subsubsection{Data Scale-up}
Following the setting of the state-of-the-art approach \cite{yang2023mask}, we progressively train our model on Human3.6M, MPI-INF-3DHP, and in-the-wild TikTok datasets, and subsequently evaluate it on the previously unseen MPII dataset. As presented in \cref{fig:scale}, our method gains improvement on scaled data and consistently outperforms \cite{yang2023mask} at each training stage, demonstrating scalability with increasing data.

For qualitative evaluation in \cref{fig:mpii}, we visualize the 3D hypotheses on novel actions with diverse poses (e.g., horse riding, boating) in MPII dataset, presenting our generalization ability on in-the-wild unseen data.

\subsubsection{Animal Pose Estimation}

Due to the shared articulation characteristics, we extend our human pose estimation pretext tasks to animal pose estimation where SMAL~\cite{Zuffi2017smal}, a parametric-driven mesh model for animals akin to SMPL for humans is utilized. Particularly, we select the well-constructed StanfordExtra dataset and an extended SMAL model~\cite{rueegg2022barc}, focusing specifically on dogs as a representative case. Despite the absence of curated distributions or textures specifically tailored to animals, our basic synthetic setting \textit{SynMH} with \textit{Seg} (in \cref{fig:texture}) is successfully employed as the animal pose estimation extension.

\begin{figure}[t]
    \centering
    \includegraphics[width=\linewidth]{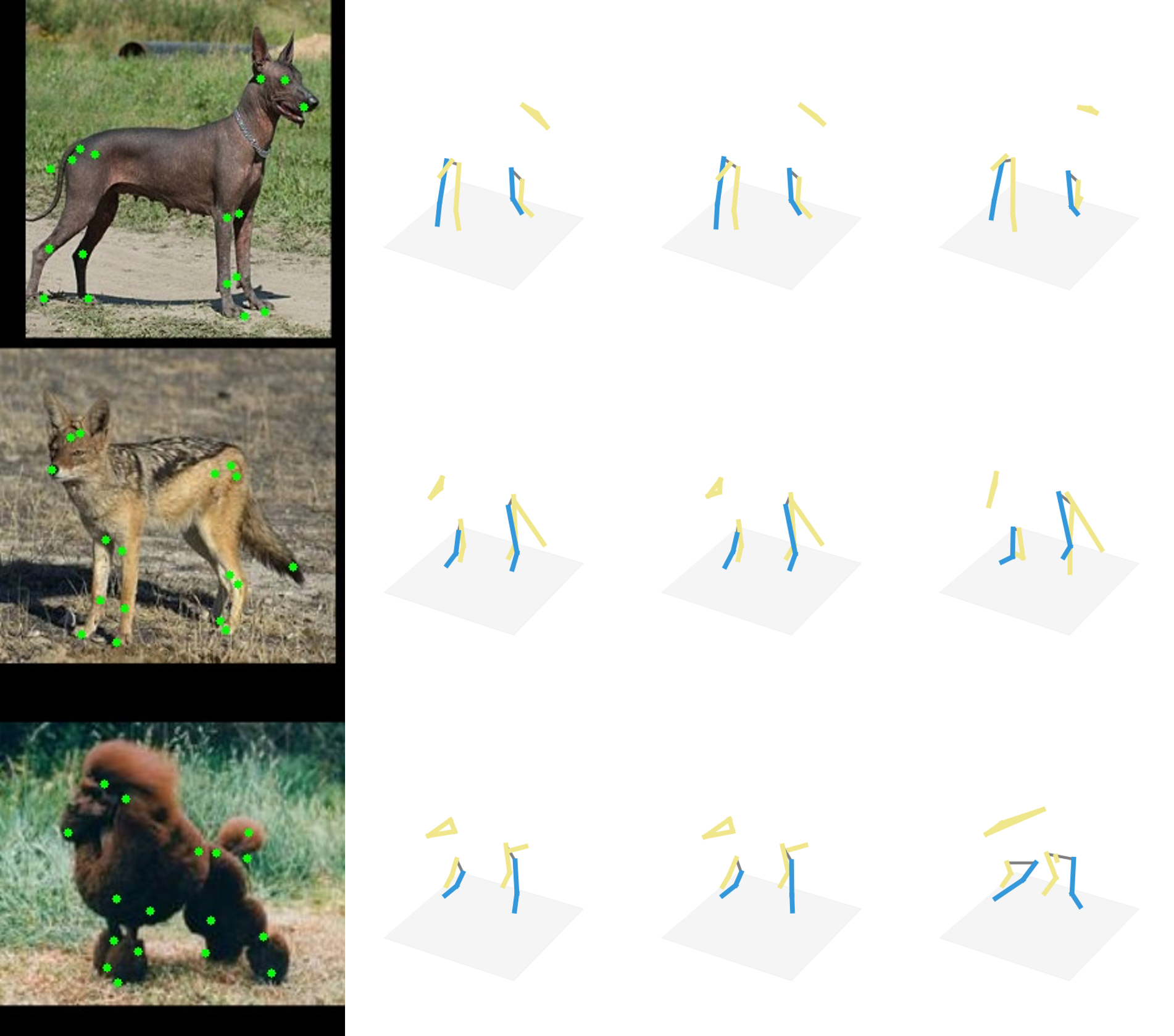}
    \caption{Visualization of the predicted 2D projected poses and 3D pose hypotheses on the StanfordExtra dataset.}
    \label{fig:dog}
\end{figure}

As demonstrated in \cref{fig:dog}, without any human-involved annotations, our method achieves plausible unsupervised monocular 3D pose estimation results, showcasing the generalization ability of our unsupervised learning concept.

\subsection{Limitations}
Apart from the discussed advantages, our approach remains limited in a few aspects. The 2D ambiguity of invisible keypoints from occlusion, a common challenge in pose estimation tasks, is not explicitly addressed. Supervised learning techniques for occlusion management could potentially be integrated.
Additionally, for concision, several existing unsupervised 3D constraints are not implemented. Given their orthogonality to the proposed framework, combining these constraints could potentially enhance performance.




\section{Conclusion}
In this paper, we discuss the challenge of depth ambiguity in unsupervised monocular 3D pose estimation. 
To mitigate this issue, we propose the X as Supervision 
framework that models the task as a multi-solution problem by a novel multi-hypothesis detector.
Correspondingly, we leverage a structured solution space based on 3D human priors from SMPL to solve it.
Specially, the proposed detector effectively maintains the inherent multimodal responses and outperforms single-hypothesis settings without adding any extra network overhead. 
The combined use of 2D mask constraints, discriminative learning, and synthetic pairs of SMPL provides robust supervision for the detector.
Experiments on the widely used datasets demonstrate superior performance.
Additionally, scalability analyses and applications to animal pose estimation demonstrate the broad generalization capacity of our unsupervised learning strategy.

\section*{Acknowlegement}
This work is partially supported by the National Key R\&D Program of China NO.2022ZD0160104.

\newpage
{
    \small
    \bibliographystyle{ieeenat_fullname}
    \bibliography{main}
}

\clearpage
\setcounter{page}{1}
\maketitlesupplementary
\appendix

\section{Multi-hypothesis Decoding in Matrix Operations}
\label{sec:supp-matrix-conversion}
To gain plausible computing efficiency, we turn the element-wise computing in the Sec. 3.2 into matrix operation. 
Specifically, the multi-hypothesis decoding involves candidate peak selection in Sec. 3.2 and local weighting for depth hypotheses in Eq. (4) operations to be optimized.

For candidate peak selection, we simply use matrix $\mathbf{H}_l^{depth} = \{\mathbf{H}_i| i=0,1..., D-3 \}$ and $\mathbf{H}_r^{depth} = \{\mathbf{H}_i| i=1..., D-1\}$ to indicate the left and right element of the depth heatmap $\mathbf{H}^{depth}$ in the range of $[1, D-2]$. Then Eq. (4) is equivalent to:
\begin{equation}
    \mathbf{p}^{cand} = \mathbf{1}(\mathbf{H}^{depth} \geq \mathbf{H}_l^{depth} \land \mathbf{H}^{depth} \geq \mathbf{H}_r^{depth}).
\end{equation}

For local weighting, the summation is equivalent to the 1D average pooling on the target dimension of a matrix and selecting the corresponding index of the window's center. Eq. (6) is computed as:
\begin{equation}
    \hat{\mathbf{z}}^{hypo}_h = \{AvgPool(\mathbf{P}\cdot \mathbf{H}^{depth}) / AvgPool(\mathbf{H}^{depth})\}(\mathbf{p}_h),
\end{equation}
where $\mathbf{P}$ indicates the index vector $\mathbf{P} = \{\mathbf{p}_i| i=0,1,...,D-1\}$ ranging from $0$ to the depth heatmap size $D$. $AvgPool$ indicates the 1D averaging pooling operation with $1$ stride, $2N_w$ neighbor size, and $N_w$ padding.

\section{GCN-based discriminator}
\label{sec:supp-gcn}
\begin{figure}[th]
    \centering
    \includegraphics[width=\linewidth]{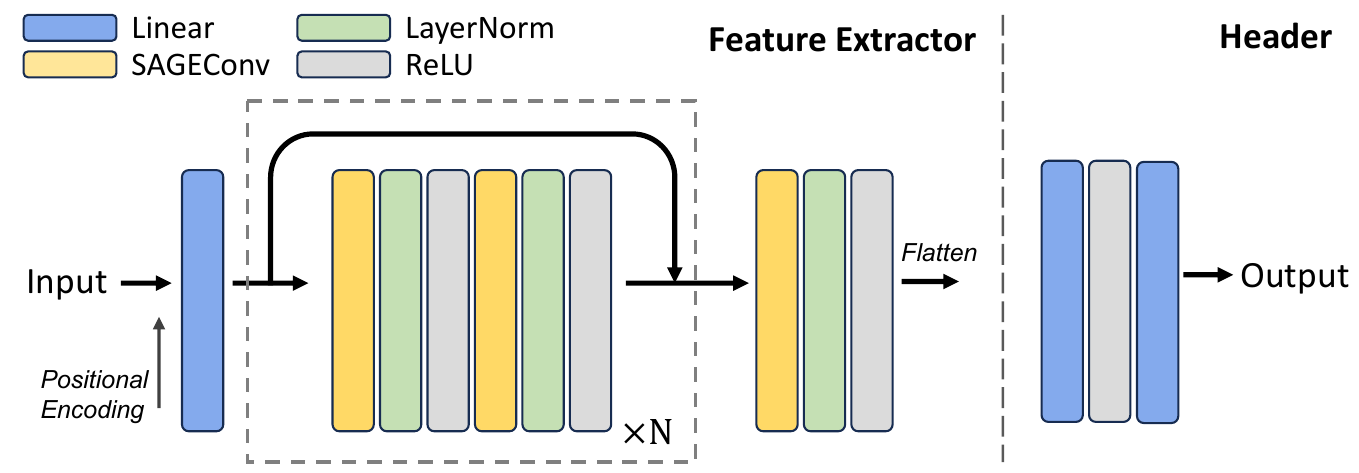}
    \caption{Network structure of GCN-based discriminator.}
    \label{fig:gcn-structure}
\end{figure}

As illustrated in \cref{fig:gcn-structure}, the discriminator follows a classical architecture of two components: the feature extractor and the header. Since the dual-representation input of keypoints and bones share the same data format, two feature extractors with identical structures are used to process the input, with positional encoding applied to each accordingly. The feature extractor starts with a Linear layer to increase the dimensionality, $N$ times GraphConv modules with two consequent SAGEConv-LayerNorm-ReLU sub-modules, alongside a residual connection. It ends with a GraphConv sub-module for refinement. The extracted features from dual-representation input are then flattened, concatenated, and passed to the header. The header, in Linear-ReLU-Linear structure, processes this merged feature to provide the final output, facilitating discriminative learning of LSGAN.

\section{Experiments}
\subsection{Implementation Details}
\label{sec:spp-implemenation}
Following \cite{sun2018integral}, we select ImageNet~\cite{deng2009imagenet} pretrained ResNet~\cite{he2016deep} as the backbone of the proposed detector. For unsupervised exploration, it keeps a similar network parameter in the existing methods and focuses on the unsupervised approach and pretext design. 
Empirically, the number of hypotheses $N_{hypo}$ is chosen as $3$ and the neighbor size $N_w$ is $15$. It fits the depth heatmap $\mathbf{H}^{depth} \in \mathbb{R}^{18 \times 64}$. The first dimension of the heatmap indicates the number of joints defined in Human3.6M~\cite{ionescu2013human3} and applied across all experiments. For the discriminator, we set the number of GraphConv modules as $2$, making it effective and simple.

For the SMPL synthetic dataset, we set a probability of $0.4$ for each pose parameter to keep the same in the original T-pose. The shape parameters are sampled from $[-1.5, 1.5]$ width of the truncated Gaussian distribution. $[45, 60], [10, 10], [30, 0], 
 [70, 0], [20, 20], [10, 10]$ in degree are the width ranges utilized for knee and hip pose parameters; $[90, 90], [50, 120], [150, 30], [60, 60], [0, 120], [15, 15]$ for shoulder and elbow pose parameters, sampling from the truncated Gaussian distribution. Spines are set in $[60, 20], [30, 30], [30, 30]$ and $[15, 15], [50, 50], [15, 15]$, respectively. The global rotation is in $[-5, 5], [-180, 180], [-5, 5]$. Following the top-down pose estimation strategy, the camera parameters are given from the training dataset for projection in camera coordinates. The synthetic dataset is pre-created in observation views with the camera parameter of Human3.6M and keeps with the same amount. For rendering, the synthetic human texture is from SURREAL~\cite{varol2017learning} and the alternative color is from randomly selecting Matplotlib colormaps by normalizing part segmentation label indexes. 

The training process is conducted in two stages, lasting 50 epochs and 15 epochs, respectively, utilizing the Adam optimizer~\cite{kingma2014adam} with batch size $32$. In the first stage, the learning rate starts at $2\times 10^{-4}$ and then decays $10$ times at epoch $40$. In the second stage, the learning rate is set to $1\times 10^{-4}$. The balancing hyper-parameters $\lambda_m$, $\lambda_g$, $\lambda_r$ are set to $2\times 10^{-2}$, $1.0$, and $0.5$ to keep the loss value at the same scale at about $10^{-4}$ in convergence.
All the experiments are conducted on NVIDIA 4090 GPUs.

\subsection{Qualitative Results}
\label{sec:supp-qualitative}
\begin{figure*}[hb]
    \centering
    \includegraphics[width=\linewidth]{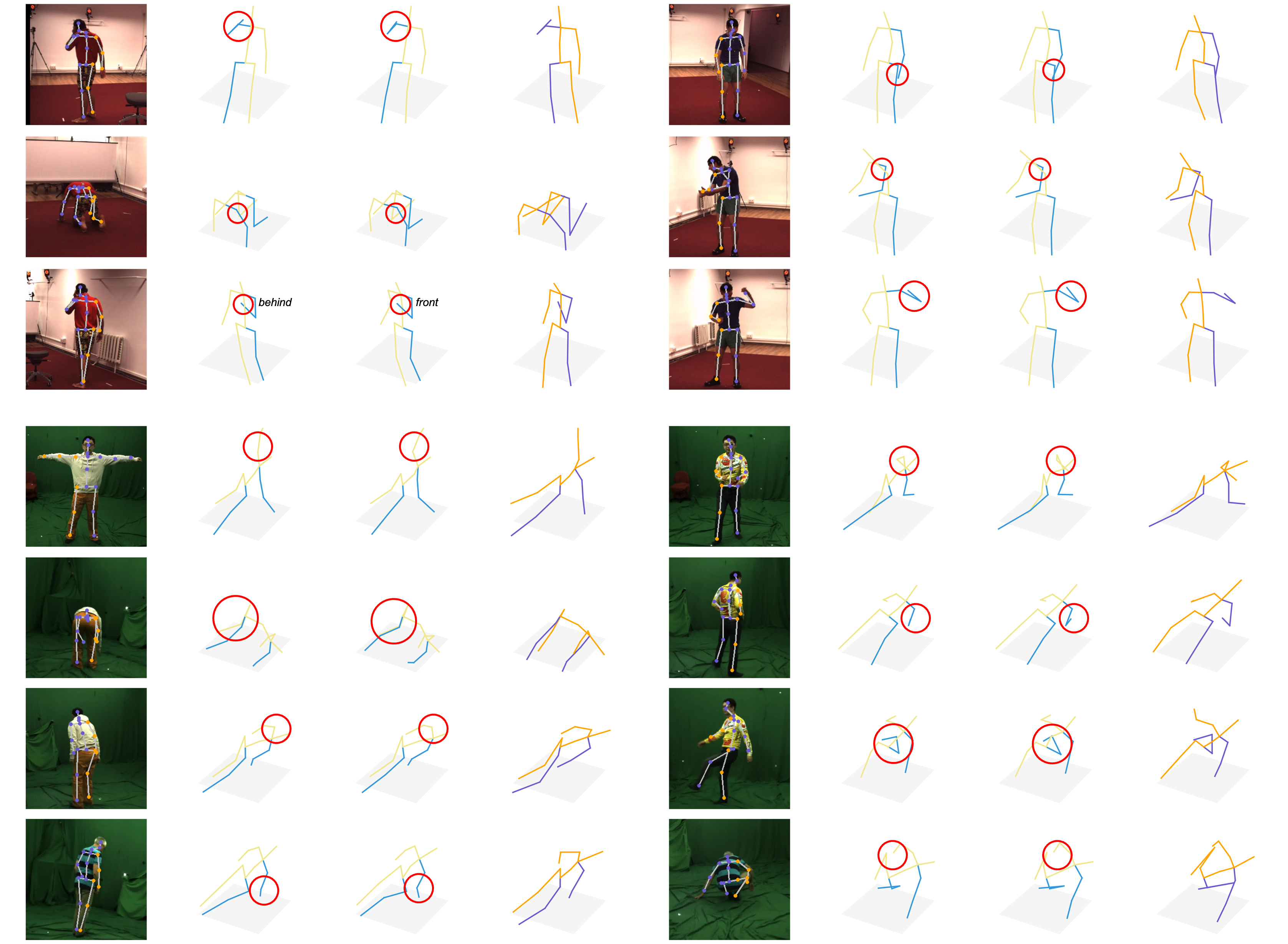}
    \caption{Qualitative results on Human3.6M (top) and MPI-INF-3DHP (bottom). From left to right in each subfigure: predictions on 2D, top two 3D hypotheses based on confidence, and ground truth 3D joints. Differences in hypotheses are highlighted in the red circles.}
    \label{fig:supp-qualify}
\end{figure*}

We visualize the approach's monocular 3D pose estimation results in \cref{fig:supp-qualify}. It demonstrates that using multi-hypothesis and integrated pretext tasks effectively captures the details of the predictions and helps alleviate ambiguity.

\end{document}